\documentclass[nohyperref]{article}

\usepackage{microtype}
\usepackage{graphicx}
\usepackage{subfigure}
\usepackage{booktabs} 

\usepackage{hyperref}

\usepackage[accepted]{icml2023}

\usepackage{amsmath}
\usepackage{amssymb}
\usepackage{mathtools}
\usepackage{amsthm}

\usepackage[acronym,symbols,nopostdot,style=super,nonumberlist,toc,section=subsection,numberedsection, hyperfirst=False]{glossaries-extra}

\usepackage[capitalize,noabbrev]{cleveref}

\usepackage{censor}
\usepackage{dsfont}
\graphicspath{{figures/}} 

\DeclareMathOperator*{\argmax}{argmax}

\DeclareMathOperator*{\mcover}{\Gamma_\Delta^{(\text{cover})}}
\DeclareMathOperator*{\mover}{\Gamma_\Delta^{(\text{over})}}

\newglossaryentry{cnn}{
    type=\acronymtype, name={CNN},
    description={Convolutional Neural Network},
    longplural={Convolutional Neural Networks},
    first={Convolutional Neural Network (CNN)}, 
    firstplural={Convolutional Neural Networks (CNNs)}
}
\newacronym{gnn}{GNN}{Graph Neural Network}
\newacronym{model}{BIC}{Radial Beam-based Image Canonicalization}
\newacronym{lstm}{LSTM}{Long Short Term Memory}
\newacronym{mlp}{MLP}{Multi-Layer Perceptron}
\newacronym{ood}{OOD}{out-of-distribution}
\newacronym{tsne}{t-SNE}{t-Distributed Stochastic Neighbor Embedding}

\theoremstyle{plain}
\newtheorem{theorem}{Theorem}[section]

\newtheorem{lemma}[theorem]{Lemma}

\theoremstyle{definition}

\theoremstyle{remark}

\newglossary[slg]{symboltype}{syi}{syg}{List of Symbols}

\newglossaryentry{its}{
    type=\acronymtype, name={ITS},
    description={Inverse Transformation Search},
    longplural={Inverse Transformation Searches},
    first={Inverse Transformation Search (ITS)}, 
    firstplural={Inverse Transformation Searches (ITSs)}
}

\usepackage[textsize=tiny]{todonotes}

\icmltitlerunning{Learning Continuous Rotation Canonicalization}

\begin{document}

\twocolumn[
\icmltitle{Learning Continuous Rotation Canonicalization \\ with Radial Beam Sampling}



\icmlsetsymbol{equal}{*}

\begin{icmlauthorlist}
\icmlauthor{Johann Schmidt}{yyy}
\icmlauthor{Sebastian Stober}{yyy}
\end{icmlauthorlist}

\icmlaffiliation{yyy}{AILab, Institute of Intelligent Cooperating Systems, Otto-von-Guericke University, Magdeburg, Germany}

\icmlcorrespondingauthor{Johann Schmidt}{johann.schmidt@ovgu.de}

\icmlkeywords{Geometric Deep Learning, Rotation Invariance, Angle Regression}

\vskip 0.3in
]



\printAffiliationsAndNotice{}  

\begin{abstract}
Nearly all state-of-the-art vision models are sensitive to image rotations.
Existing methods often compensate for missing inductive biases by using augmented training data to learn pseudo-invariances.
Alongside the resource-demanding data inflation process, predictions are often poorly generalised.
The inductive biases inherent to convolutional neural networks allow translation equivariance through kernels acting parallel to the horizontal and vertical axes of the pixel grid.
This inductive bias, however, does not allow for rotation equivariance.
We propose a radial beam sampling strategy along with radial kernels operating on these beams to incorporate centre-rotation covariance.
We present a radial beam-based image canonicalisation (BIC) model with an angle distance loss.
Our model allows for continuous angle regression and canonicalises random centre-rotated input images.
As a pre-processing method, this enables rotation-invariant vision pipelines with model-agnostic rotation-sensitive downstream predictions.
We show that our angle regressor can predict continuous rotation angles and improves downstream performances on \textit{COIL100}, \textit{LFW} and \textit{siscore}.
Our code is publicly available at \url{github.com/johSchm/RadialBeams}. 
\end{abstract}

\section{Introduction}
\label{sec:introduction}

\begin{figure}[ht]
\vskip 0.2in
\begin{center}
\centerline{\includegraphics[width=\columnwidth]{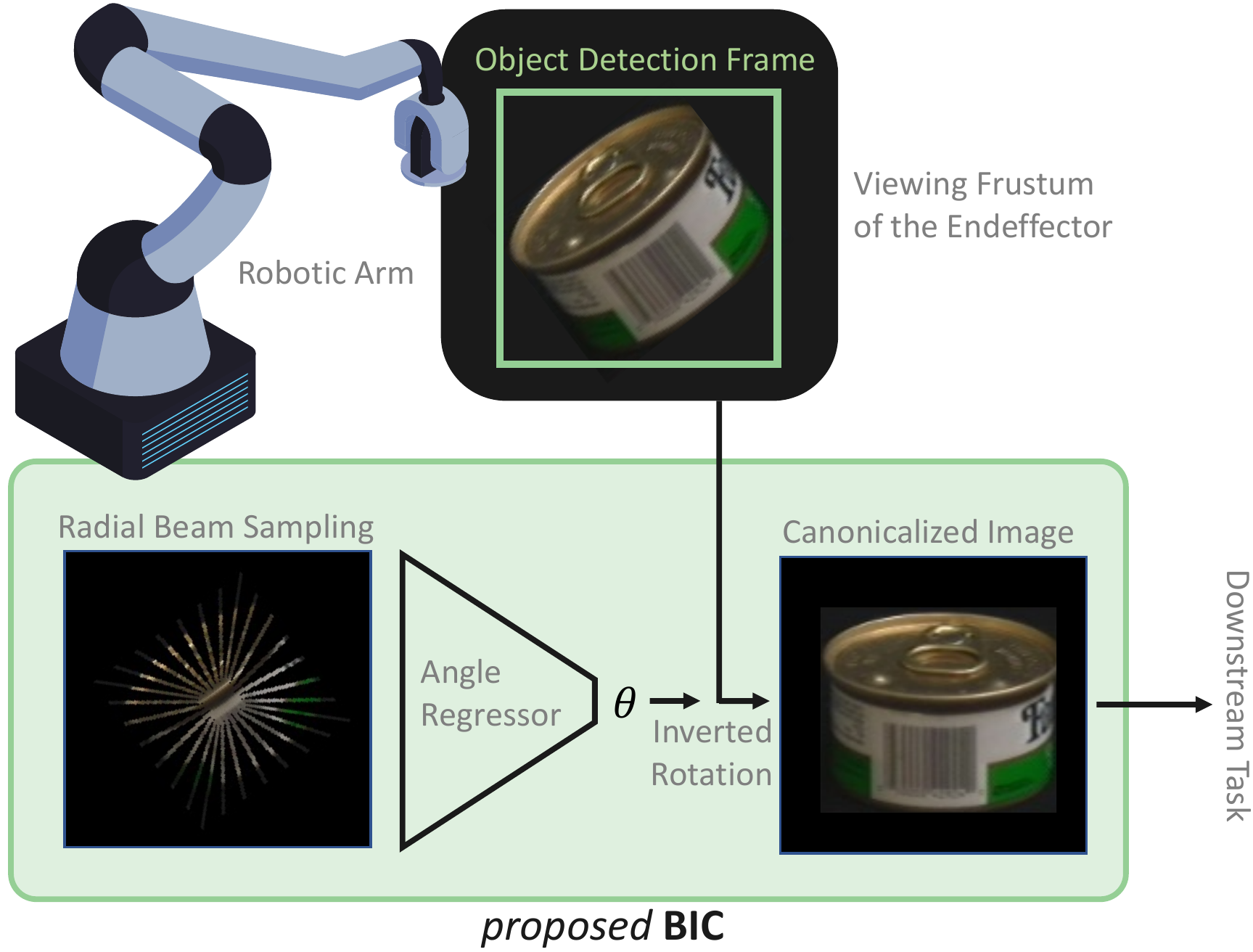}}
\caption{Industrial \glsxtrfull*{model} application setting.}
\label{fig:highlevel}
\end{center}
\vskip -0.2in
\end{figure}


Computers often struggle to detect the high-level semantic similarity between rotated objects, a task trivial to humans.
During the rotation transformation, the object's structure (semantic similarity) is preserved. 
We say the rotation is a symmetry of that object.
Defying these symmetries would confront us with abundant semantically redundant information.
The human brain learns to discard these symmetry transformations to free its immanent representations from these redundancies, as shown by \cite{Leibo2011}.
From an Information Theoretical point of view, avoiding redundancies allows for more efficient usage of computing capacities.
With this at hand, we can argue from a human cognition and mathematical perspective for an increased leverage of symmetries in novel deep learning architectures.
A logical first step involves integrating \textit{a priori} known symmetries in our artificial systems.
In this work, we narrow down the scope to image processing. 
\glspl*{cnn} revolutionised the computer vision field and now are indispensable in modern deep learning and its applications \cite{HAR21}.
This is grounded on the reduced parameter footprint through shared filter parameters and spatially-independent feature detection, which introduced translation equivariance to \glspl*{cnn} (see \cite{cohen2016group} for a proof).
Although translation-equivariant architectures are a major step toward efficient vision models, many known visual symmetries are disregarded.
In this work, we aim to extend the symmetry scope of vision pipelines by rotation-invariance.
We focus on image classification problems, where arbitrary image rotations keep the semantic label unchanged (see the \textit{siscore} dataset \cite{djolonga2020robustness}).
Generally, three options exist for how to integrate in-/equivariances in deep learning models:
\\
\textit{(i)} The integration of an inductive bias to render the architecture inherently invariant/equivariant to the target symmetry.
Equivariance limits the function space to valid models under the rules of natural image formation \cite{Worrall2017}, which supports learning and generalisation.
For instance, attempts by \cite{cohen2016group, zhou2021metalearning} and \cite{ecker2018rotationequivariant} to integrate rotation-equivariance in \glspl*{cnn} are non-trivial, come with decisive constraints in the filter space, and are limited to discrete rotation subgroups.
\\
\textit{(ii)} The augmentation of the training set inflates the dataset by rotated versions of the vanilla image \cite{Chen2019InvarianceRV}.
These augmented examples are sampled from the symmetry orbits of the vanilla images (see \cref{sec:symmetry}).
During training, the network learns to approximate these orbits on the latent manifold (pseudo-equivariance) or collapse them (pseudo-invariance).
While straightforward, this significantly increases the required sample complexity to sufficiently cover the symmetry space.
However, due to the lack of alternatives, the vast majority of modern vision models employed in an industrial environment still rely purely on augmentation \cite{Saeed2021}.
In \cref{sec:related_work}, we pursue a more profound investigation of the related work.
\\
\textit{(iii)} Alternatively, a surjective projection during pre-processing can be used to collapse the orbit of an image to its canonical form.
This canonicalisation process is employed in the human brain, where canonical orientations of different objects are stored, as shown by \cite{Harris2001}.
In this work, we investigate the technical realisation of this process using deep learning.
We design a continuous angle regressor to estimate the rotation angle of input images.
This should answer the question: \textit{Given an arbitrary centre-rotated image, what is its canonicalised orientation?}
The model learns image canonicalisation and maps all possible rotations of an image, \textit{i.e.}, the image orbit, to a single orientation.
In practice, our model is placed between an object detector and a rotation-sensitive downstream model, as illustrated in \cref{fig:highlevel}.
Object detection acts as an additional symmetry reducer by pruning all but the relevant pixel information.
This also removes the centre bias posed by the proposed radial beam sampling, which will be introduced in  \cref{sec:methodology}.
Our model does not pose any restriction in terms of the downstream task or model used.
All this is based on radial beam sampling, where beams of pixels originating from the centre of the image to its edge region.
This introduces a covariance between center-rotations of the underlying image and circular shifts of the sampled beams.
We propose an inductive bias, which integrates this covariance into the encoder of our proposed \gls*{model} model.
In contrast to \cite{Keller2021TopographicVL}, this covariance does not need to be learned.
Our model approximates the entire rotational Lie group, rather than limited discrete rotation subgroups as of prior works, like \cite{ecker2018rotationequivariant, Ustyuzhaninov2020Rotation}.
Compared to rotation-equivariant inductive biases as in \cite{Gens2014}, our approach does not bind the hypothesis class and simultaneously lowers the sample complexity of the downstream model.
We provide more methodological details in \cref{sec:methodology}, which we thoroughly evaluate in \cref{sec:experiments}.

\section{Symmetries and Group Theory}
\label{sec:symmetry}

Mathematically, symmetries of an object are comprised in a group, \textit{i.e.}, a set with an operation.
A symmetry of an object is a transformation that preserves the object's structure, like the $120^\circ$ rotation of an equilateral triangle.
Therefore, the rotations, which preserve the structure of the triangle, form a discrete group containing $0^\circ, 120^\circ$ and $240^\circ$.
Besides discrete symmetry groups, Lie groups are continuous symmetry groups whose elements form a smooth differentiable manifold, like the rotation group of a circle.
Of uttermost interest in this paper are two-dimensional rotations described by the Special Orthogonal Group $\texttt{SO(2)}$.
This group comprises distance-preserving transformations of the Euclidean space, which is an essential property for image rotations.
Formally, the group is defined by
\begin{equation}
  \label{eq:so2}
  \texttt{SO(2)} = \left\{ \mathfrak{g} : \rho(\mathfrak{g}) = R_\theta \equiv \left[\begin{array}{cc}
    \cos(\theta) & -\sin(\theta) \\
    \sin(\theta) & \cos(\theta)
  \end{array}\right] \right\}, 
\end{equation}
where the rotation angle $\theta \in [0^\circ, 360^\circ)$ is expressed by the group element $\mathfrak{g}$ using the homomorphism between $\texttt{SO(2)}$ and angles $\theta$ as in \cite{hall2015lie}.
$\rho(\mathfrak{g})$ is called the representation of $\mathfrak{g}$, \textit{i.e.}, the rotation matrix.
The group comes with the group operation $+_{360^\circ}$, such that, for instance, $220^\circ +_{360^\circ} 150^\circ = 10^\circ$.
For an arbitrary group $\mathfrak{G}$, we define the orbit $\text{orb}(\cdot)$ as the set of all images under the group action with respect to a fixed element $x$, \textit{i.e.}, $\text{orb}(x, \mathfrak{G}) = \{ \rho(\mathfrak{g})x \mid \mathfrak{g} \in \mathfrak{G} \}$.
A function $f$ (\textit{e.g.}, modelled by a neural network), which follows the symmetry group $\mathfrak{G}$, is either equivariant, invariant, or covariant \cite{Marcos_2017}.
Out of these three categories, covariance is the most general one, where $f(\rho(\mathfrak{g}) x) = \rho^{\prime}(\mathfrak{g}) f(x), \forall \mathfrak{g} \in \mathfrak{G}$.
Both transformations might be potentially different in their domain and co-domain, respectively.
From this definition, we can derive the following two special cases:
If the output stays unaltered after transforming the input, $f(\rho(\mathfrak{g}) x) = f(x), \forall \mathfrak{g} \in \mathfrak{G}$, we call $f$ invariant.
If the output is transformed equally, $f(\rho(\mathfrak{g}) x) = \rho(\mathfrak{g}) f(x), \forall \mathfrak{g} \in \mathfrak{G}$, we call $f$ equivariant.
For more details please refer to \cite{bronstein2021geometric}.
In this work, we strive after continuous rotation invariances following the Lie group $\texttt{SO(2)}$.
However, as shown in the related work, most rotation symmetric vision networks respect only narrow subgroups of $\texttt{SO(2)}$.

\section{Related Work}
\label{sec:related_work}

Through its simplicity, the process of augmented training is a prevalent strategy for learning more robust models.
The dataset is augmented with samples from the loss-preserving symmetry orbit of a datapoint, like rotations of a base image.
During training, pseudo-invariances should be embedded into the latent manifold by learning proximity regions of pseudo-invariant data points in the latent space. \cite{Goodfellow2009} show that the quality of pseudo-invariance in the learned representation increases with the depth of the network. Minor deformations and transformations, like shifts and rotations, are attenuated by lower layers, while higher layers progressively form invariances.
As shown by \cite{Chen2019InvarianceRV} and \cite{kernelDA2019}, augmentation acts as a regulariser penalising model complexity by learning to minimise the average over loss signals under the group actions.
Despite these benefits, modern networks depend on massive training datasets full of semantic redundancies, whose sample complexity is even enlarged by additional augmentation processes.
\\
Contrary to hard-won pseudo-invariances, inductive biases open the gates for true invariances \cite{Nabarro2021DataAI}.
\cite{Gens2014} leverage the aforementioned proximity regions of invariant datapoints by enforcing layers to model the symmetry space of inputs deliberately.
Hence, points propagated through the network can be pooled with their closest neighbours (\textit{i.e.}, their symmetry transformed replicates) and mapped to the same output label. \cite{jaderberg2016spatial} ingrain jointly learned transformations to \glspl*{cnn}, which project inputs to their canonical representations.
The end-to-end trained transformer is limited to small perturbations of inputs, and convergence to satisfactory local minima is challenging.
Both limitations are tackled by our \gls*{model} model.
\cite{hu2019exploring} constrain filters to symmetric weight matrices, \textit{i.e.}, isotropic filters. 
This also reduces the memory footprint since only a sub-matrix must be stored.
\cite{cohen2016group} expand the symmetry range of \glspl*{cnn} to reflections and integer multiples of $90^\circ$ rotations around the centre.
\cite{zhou2021metalearning} decompose weight matrices into learnable weight vectors and meta-learned (binary) symmetry matrices. 
\cite{ecker2018rotationequivariant} represent kernels in a steerable basis modelled by 2D Hermite polynomials to allow for weight sharing across kernel orientations.
Building upon these results, \cite{Ustyuzhaninov2020Rotation} canonicalise the kernel weights for different orientations.
Most proposed methods drastically constrain the filter space, making arbitrary visual feature learning impossible.
Furthermore, due to the finite number of kernel matrices and rotational artefacts, only a limited number of discrete rotations is supported.
Both drawbacks are eliminated with our proposed method.

\section{Methodology}
\label{sec:methodology}

\begin{figure}[ht]
\vskip 0.2in
\begin{center}
\centerline{\includegraphics[width=0.8\columnwidth]{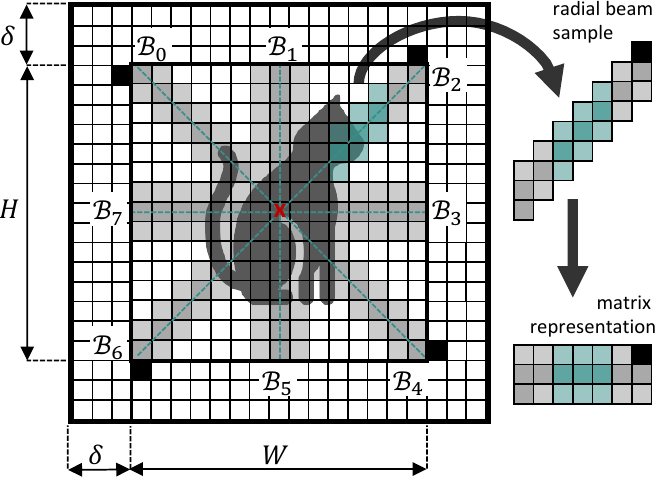}}
\caption{Beam set $\mathcal{B}$ on the padded image $X$ with beams (grey), proximity/thickness $\epsilon$ (light grey), padded pixels (black) and the radial convolutional kernel (cyan).}
\label{kernel}
\end{center}
\vskip -0.2in
\end{figure}


\begin{figure*}[ht]
\vskip 0.2in
\begin{center}
\centerline{\includegraphics[width=\textwidth]{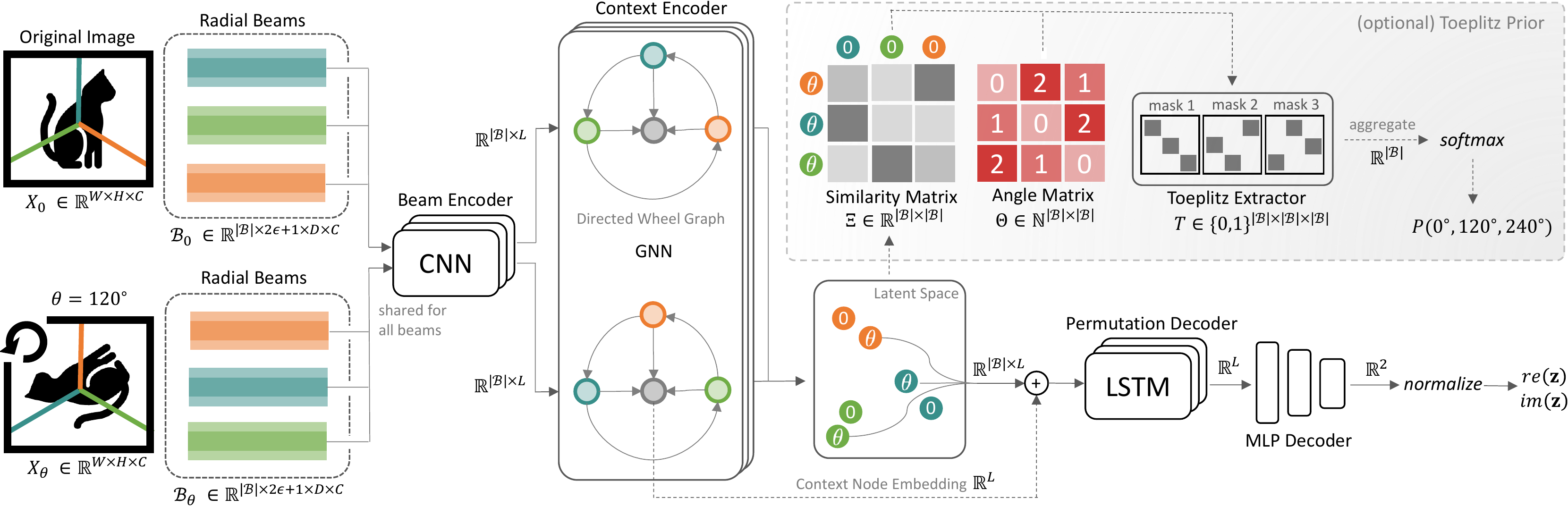}}
\caption{A schematic illustration of our angle regressor.
  The proposed Toeplitz prior for training utilizes a tuple containing image $X_0$ and its $\theta$-rotated version, $X_\theta$.
  First, radial beams $\mathcal{B}_0$ and $\mathcal{B}_\theta$ are sampled with proximity and embedded by a shared \gls*{cnn} encoder. A \gls*{gnn} encodes the neighbourhood information of adjacent beams.
  During training time, we impose a prior on the resulting latent manifold to improve its structure.
  To that end, a pairwise similarity matrix $\Xi$ over embeddings of $\mathcal{B}_0$ and $\mathcal{B}_\theta$ is computed.
  We show that $\theta$ is mapped to diagonals in $\Xi$.
  Leveraging this insight, we can extract the probability distribution over discrete rotations from $\Xi$ by using a Toeplitz extractor $T$ conditioned on the angle matrix $\Theta$.
  During inference, however, we are limited to a single input image $X_\theta$.
  To find the rotation angle, we leverage the permutation of beam embeddings using a \gls*{lstm} decoder.
  We use a complex unit vector regression by jointly predicting the real and imaginary parts.}
\label{fig:model}
\end{center}
\vskip -0.2in
\end{figure*}

We present an orientation canonicaliser for image data.
At its heart, our model uses radial beams sampled from the input image as illustrated in \cref{fig:kernel}.
This is followed by an angle regressor as shown in \cref{fig:model}.
The predicted angle is used to canonicalize the input image using the inverted rotation.
Let $X \in \mathcal{X}(\Omega)$ be an image on the pixel grid domain $\Omega$ sampled from the dataset $\mathcal{X}$.
We have an image tensor $X \in \mathbb{R}^{W \times H \times C}$ with width $W$, height $H$, and color channels $C$.
In practice, we use batched data, but we drop the batch dimension in this article for brevity.

\paragraph{Radial Beam Sampling}


Conventional convolution on image data uses horizontally and vertically aligned rectangular kernels \textit{w.r.t.} $\Omega$.
This allows for translation-equivariance when shifting the object along either of these axes.
In this work, we are interested in centre-rotations of images rather than translations.
We argue that one way to enable rotation-equivariance is to alter the alignment of kernels.
To this end, we utilize kernels operating along beams radiating from the centre to the edges of the image.
Formally, a beam $\boldsymbol{b}$ is a fixed vector radiating from $\left[ \left\lfloor \frac{W}{2} \right\rceil, \left\lfloor \frac{H}{2} \right\rceil \right]$ with length $D$ and a certain direction.
We use a deterministic sampling strategy to obtain a radial beam set $\mathcal{B}$, as illustrated in \cref{fig:kernel}.
Since this is a rigid overlay mask on top of $\Omega$, beam positions are agnostic to any image $X$.
This allows us to check for intersections of $\mathcal{B}$ with the pixel grid $\Omega$ once.
Afterwards, we only need to extract the colour information of image $X$ at these positions to evaluate the beams.
In practice, this can be implemented via the \cite{Bresenham1965AlgorithmFC} Algorithm.
The cardinality $|\mathcal{B}|$ is a hyper-parameter, which is upper-bounded by $|\mathcal{B}_{max}|$ (see \cref{lemma:beam_upper_bound}).
Beams are spaced equally from one another to obtain a near uniform receptive field and reduce bias in the sampling.
We control the thickness of beams by $\epsilon \in \mathbb{N}$.
To this end, we modify the classical Bresemham Algorithm and add additional neighbourhood pixels, such that with $\epsilon = 1$, the sampled line is enlarged by a border of one pixel to both sides, increasing the width to $2\epsilon + 1$.
To further understand the information density extracted from $X$ by $\mathcal{B}$, we propose an approximation of the beam coverage and the resulting overlaps \textit{w.r.t.} $|\mathcal{B}|$ and length $D$.

\begin{lemma}
\label{lemma:cover}
  The pixel coverage $\Gamma_{cover}$ of radial beams sampled from image $X$ is given by
  \begin{equation}
    \Gamma_{cover}(|\mathcal{B}|, D) \approx
    |\mathcal{B}| \left[ \frac{4 D^2}{|\mathcal{B}|} - \frac{\left( \frac{8D}{|\mathcal{B}|} - (2\epsilon + 1) \right)^2}{2\tan \left(\frac{360^\circ}{|\mathcal{B}|} \right)} \right].
  \end{equation}
  Then, the pixel overlap $\Gamma_{over}$ can be estimated by
  \begin{align}
    \Gamma_{\text {over }}(|\mathcal{B}|, D) \approx|\mathcal{B}| & \Biggl[ { (2 \epsilon+1) D-\frac{4 D^2}{|\mathcal{B}|} } \nonumber \\
    & +\frac{\left(\frac{8 D}{|\mathcal{B}|}-(2 \epsilon+1)\right)^2}{2 \tan \left(\frac{360^{\circ}}{|\mathcal{B}|}\right)} \Biggl]
    \end{align}

\end{lemma}
\begin{lemma}
  \label{lemma:beam_upper_bound}
  The maximum number of beams is upper bounded by $|\mathcal{B}_{max}| \leq 8D / (2\epsilon+1)$.
\end{lemma}
The proofs are stated in  \cref{app:overlap} and \cref{app:beam_upper_bound}, respectively.

\paragraph{Shared Beam Encoder}
\label{sec:thickness}
As outlined in the following, we embed the sampled beams using a shared beam encoder to mitigate coarsening issues.
Through the trigonometrical functions $\sin$ and $\cos$ in \cref{eq:so2}, the rotation matrix $R$ can take real values even for integer angles.
Mapping the rotated pixel positions back to the coarse pixel grid introduces two problems:
\textit{(i)} Pixels might end up outside the pixel grid, where the corners of images are at risk.
This can be circumvented by isotropic zero-padding and thus increasing the size to $W + 2 \delta$ and $H + 2 \delta$.
\begin{lemma} \label{lemma:padding}
  The optimal padding $\delta$ for any squared image $X$, which obviates any information loss when rotated under group transforms of $\texttt{SO(2)}$, is given by $\delta = \max \left(0, \left\lceil W \left(\sqrt{2} - 1 \right) / 2 \right\rceil \right)$.
\end{lemma}
The proof is stated in  \cref{app:padding}.
Note that the padding does not affect the prediction and is only used to prevent information loss during group transformations.
\textit{(ii)} Rotations also introduce rounding errors on $\Omega$.
Due to the real matrix $R$ and the discrete pixel grid, an integer constraint must be enforced after rotation.
Thus, destination locations of pixels might be allocated multiple times where others are missed entirely.
We smoothen the rotation by bilinear interpolation between pixels to mitigate these artefacts.
For further mitigation, we leverage the proximity around each pixel. 
The beam encoder embeds the proximity-aided beam evaluations in a latent representation, \textit{i.e.}, $\boldsymbol{b} \in \mathbb{R}^{2\epsilon + 1 \times D} \rightarrow \mathbb{R}^{L}$, where $L$ is the latent space dimensionality.

\paragraph{Context Encoder}
The representation emerging from the beam encoder ties spatial information together with colour values.
Since this is done individually for each beam, we mitigate greedy decisions by incorporating the spatial context through a \acrfull*{gnn}.
See \cref{app:gnns} for more details on \glspl*{gnn}.
A directed wheel graph models the context encoder with $|\mathcal{B}| + 1$ nodes.
The graph topology encodes two essential properties of information exchange during representation learning.
First, we leverage direct links to the centre node of the wheel graph to aggregate messages from all beam nodes.
Therefore, the centre node representation will hold the full global context.
Secondly, we use directed links over undirected links to avoid permutation invariant updates.
The spatial information of whether an adjacent node is above or below is crucial to quantify rotations.
This also allows us to mitigate over-smoothing \cite{Li2018Oversmooth} by holding back the global context and adding it after the \gls*{gnn}.
In \cref{sec:implementation_details}, we give more technical details, for instance, regarding the message passing strategy.


\paragraph{Toeplitz Prior}
The Toeplitz Prior represents an optional analytical aid to improve the latent structure.
We aim to predict the rotation angle given the beam representations obtained after both encoding steps.
To this end, we utilize the input tuple $(X_0, X_\theta)$ and learn clusters of similar beam embeddings, where $X_\theta = R_\theta X_0$ with $R_\theta \in \mathcal{C}_{|\mathcal{B}|}$. $\mathcal{C}_{|\mathcal{B}|}$ is a subgroup of $\texttt{SO(2)}$,
\begin{equation} \label{eq:sym_group}
  \mathcal{C}_{|\mathcal{B}|} = \left\{ R_\theta : \theta = \frac{k}{|\mathcal{B}|} 360^\circ \mid k \in \left\{0,1, \ldots,|\mathcal{B}|-1 \right\} \right\}.
\end{equation}
In other words, we rotate images during training only by angles between two beams (not necessarily adjacent ones).
We encode $\mathcal{B}$ as a tensor, hence each beam $\boldsymbol{b} \in \mathcal{B}$ has a certain position in $\mathcal{B}$.
Without loss of generality, we set the first beam as the upper left one.
In this light, rotating by $R_\theta \in \mathcal{C}_{|\mathcal{B}|}$ will shift/roll the positions of beams in $\mathcal{B}$.
This introduces a bijection between $k$ from \cref{eq:sym_group} and $\theta$, such that rotations in the ambient space map to circular shifted beam embeddings.
Hence, we need to quantify how many positions the beams are shifted to infer the image rotation.
For instance, in \cref{fig:model} we rotated the image by $\theta = 120^\circ$ and sampled $|\mathcal{B}|=3$.
According to \cref{eq:sym_group} we have $k = (\theta |\mathcal{B}|) / 360^\circ = 1$, so the rotation is mapped to a shift of beams, as illustrated by the beam order in \cref{fig:model}.
To find $k$, matching beam pairs of both images need to be identified (illustrated by equal colours in \cref{fig:model}).
We compute a similarity matrix $\Xi$ using all pairwise beam combinations with $1/(1 + ||\cdot||_2)$ as the similarity measure.
The hypothesis is that the similarity is highest for beam pairs with the target angle $\theta$ between them.
As the Toeplitz Prior in \cref{fig:model} shows, this accumulated in a shifted diagonal in $\Xi$. 
We say diagonals in $\Xi$ are coherent to circular shifts of beams in $\mathcal{B}$.
For each rotation in the finite group $\mathcal{C}_{|\mathcal{B}|}$, we have such a diagonal, which is formulated in the angle matrix $\Theta$.
We can quantify the rotation angle $\theta$ by finding the maximally activated diagonal.
For this, we mask each diagonal by the Toeplitz extraction tensor $T$, a binary mask for matrix diagonals.
By summing up all similarity scores we obtain the final logit score for $k$, \textit{i.e}, $\mathds{1}^\top (\Xi \odot T_k) \mathds{1}$, where $\mathds{1}$ is a one-vector of length $|\mathcal{B}|$ and $\odot$ denotes the Hadamard product.
Finally, we estimate the probability distribution over angles by applying the softmax function over the logits. 
Due to the \textit{a priori} defined $T$ this prior is fully differentiable.

\paragraph{Decoding}
The Toeplitz prior enforces a deliberate structure on the latent manifold.
This structure needs to be leveraged during inference to determine rotation angles of single input images $X_\theta$ without $X_0$.
We enforce order on $\mathcal{B}$ by the encoding as a tensor.
We build upon the assumption that during training, a structured latent manifold is learned on which the order of context-aware beams infers rotation information.
During inference, we utilize this permutation of $X_\theta$ and decode the sequence of beams using a \acrfull*{lstm}.
This is followed by linear transformations, such that the decoder maps $\mathbb{R}^{|\mathcal{B}| \times L} \rightarrow \mathbb{R}^{L} \rightarrow \mathbb{R}^{2}$.

\paragraph{Unit Circle Loss}
Instead of predicting a real number for $\theta$, we leverage the specific structure of angles.
Let $\boldsymbol{z}$ be a complex vector on the unit circle.
As stated in complex analysis, the angle between the positive real axis and $\boldsymbol{z}$ is $\arg{\boldsymbol{z}} = \theta$, such that each $\boldsymbol{z}$ maps to a $\theta$.
The argument of $\boldsymbol{z}$ can be computed as $\arg{(\boldsymbol{z})} = \text{atan2}(im(\boldsymbol{z}), re(\boldsymbol{z}))$, where $im(\boldsymbol{z})$ and $re(\boldsymbol{z})$ are the imaginary and real part of $\boldsymbol{z}$, respectively.
Predicting $\boldsymbol{z}$ will bound the predicted angle to $[0^\circ, \text{ } 360^\circ)$ and introduce smooth transitions from $359^\circ$ to $0^\circ$.
In practice, we normalize the output of the final linear layers to the unit length $\boldsymbol{z} / ||\boldsymbol{z}||_2$.
A loss is required which penalizes the predicted unit vector $\boldsymbol{z}$ \textit{w.r.t.} its ground truth angle $\theta$.
We use the squared error between the real and imaginary parts,
\begin{equation} \label{eq:circle_loss}
  \mathcal{L}_{\text{circle}}(\theta, \boldsymbol{z}) = (\sin(\theta) - im(\boldsymbol{z}))^2 + (\cos(\theta) - re(\boldsymbol{z}))^2.
\end{equation}

The unit circle loss is a highly expressive loss function, as it preserves angle distances (\cref{app:circle_loss}).
We provide bounds in \cref{app:extrema} and show its Lipschitz smoothness in \cref{app:lipschitz}.
The Toeplitz prior acts via a supplementary loss term,
\begin{equation} \label{eq:loss}
  \mathcal{L}(\theta, \boldsymbol{z}, P) =
  \mathcal{L}_{\text{circle}}(\theta, \boldsymbol{z}) + \mathcal{L}_{\text{prior}}(\theta, P)
\end{equation}
and 
\begin{equation} \label{eq:loss}
    \mathcal{L}_{\text{prior}}(\theta, P) = \sum_{i=0}^{|\mathcal{B}|-1} P^{(i)}_\theta \log P^{(i)},
\end{equation}
where $P_\theta$ is a one-hot vector of length $|\mathcal{B}|$ that indicates the true rotation angle $\theta$ and $P$ is the predicted distribution over angles.
\section{Experiments}
\label{sec:experiments}

\begin{figure*}[ht]
\vskip 0.2in
\begin{center}
\centerline{\includegraphics[width=\textwidth]{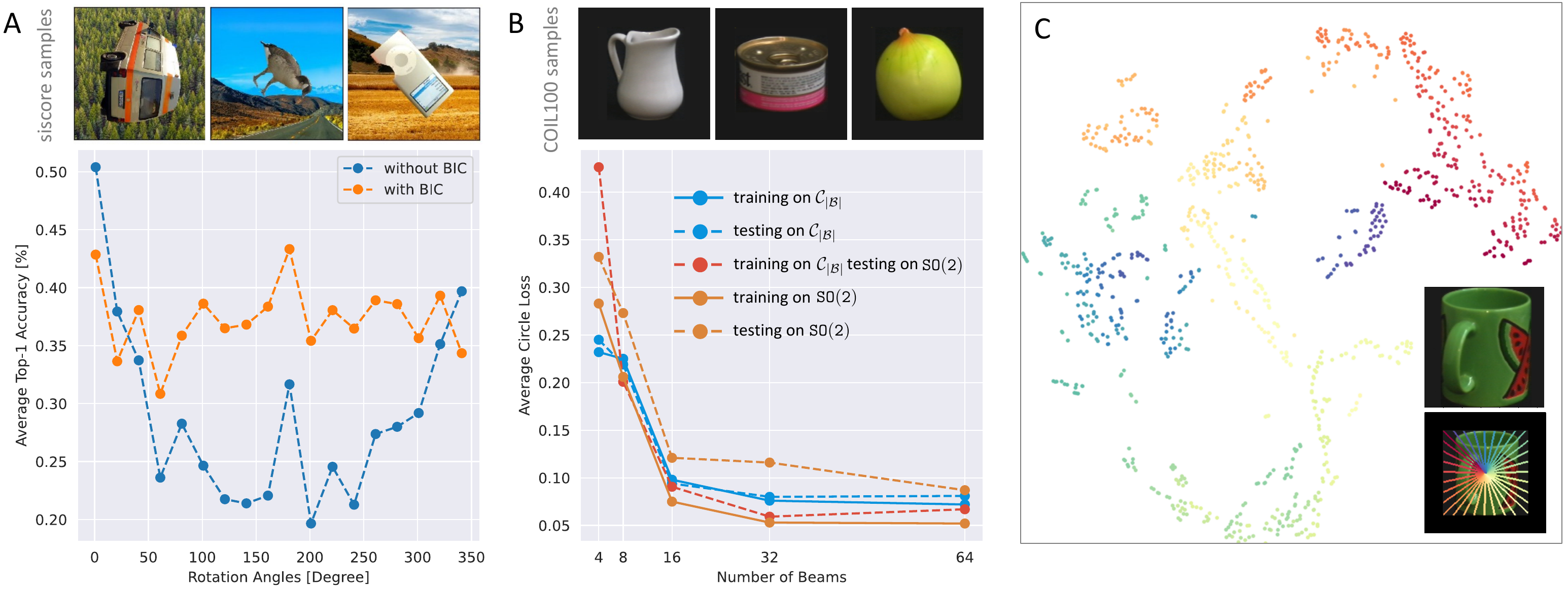}}
\caption{\textit{(A)} Performance gain and stabilization across rotations. 
\textit{(B)} Performance comparison with varying $|\mathcal{B}|$. We train/test on either the finite rotation group, $\mathcal{C}_{|\mathcal{B}|}$, or on all possible rotations, $\mathtt{SO(2)}$. Continuous lines represent training curves, and dashed lines represent testing curves. 
\textit{(C)} \gls*{tsne} projection of beam embeddings ($\mathcal{C}_{|\mathcal{B}|}$ orbit).}
\label{fig:experiment_plot}
\end{center}
\vskip -0.2in
\end{figure*}

Unless stated explicitly, we used the settings outlined in \cref{sec:implementation_details}.
In \cref{sec:downstream} we tested the robustness of \gls*{model} to different backgrounds.
In \cref{sec:structure} we showed that the representations learned indeed form a continuous orbit in the latent space.
In \cref{sec:receptive_field} we studied the generalization capabilities of \gls*{model} under varying $|\mathcal{B}|$ and found that \gls*{model} is able to fit $\mathtt{SO(2)}$. 
In \cref{sec:light} we showed that \gls*{model} leverages light reflections as a key feature.
In \cref{app:beam_lengths} we provide an investigation of the influence of different beam lengths.


\subsection{Downstream Performance Gain} \label{sec:downstream}

We evaluated our model on the benchmark dataset \textit{siscore} \cite{djolonga2020robustness}.
This benchmark for modelling robustness to Euclidean group transformations contains image with varying backgrounds and group-transformed \textit{Imagenet} \cite{deng2009imagenet} class objects. 
We used a pre-trained \textit{EfficientNetB0} by \cite{efficientnet} on \textit{Imagenet} as a classifier for \textit{siscore}.
We train \gls*{model} on $80\%$ of \textit{siscore} to learn the rotation-canonicalisation and leverage the remaining $20\%$ for evaluating the classification performance.
During training the Toeplitz prior was disabled for higher flexibility on $|\mathcal{B}|$.
Thus, the model performed a continuous unit vector regression for $\texttt{SO(2)}$.
We hypothesise that if \gls*{model} learns to canonicalise images in \textit{siscore}, the downstream classifier will benefit from the canonicalised inputs, causing a performance increase with more robustness against rotations.
Our results are shown in \cref{fig:experiment_plot}A.
\gls*{model} lifts the top-1 accuracy on \textit{siscore} to a stable level of over $0.37$ on average.
Since the learned canonicalisation is not optimal (see \cref{app:downstream} for details), the performance with \gls*{model} across rotations is not as high as for non-rotated inputs.

\subsection{Learning Continuous Representation Orbits}
\label{sec:structure}


The latent structure should also accompany our hypothesis that image rotations cohere with circular shifts of beam embeddings.
We test this by computing the beam embeddings of the entire group orbit of $\mathcal{C}_{|\mathcal{B}|}$.
The beam embeddings should form a surface topologically equivalent to a circle.
\cref{fig:experiment_plot}C shows the 2D \acrfull*{tsne} \cite{tsne} projection of a all beam embeddings of \textit{COIL100} sample.
We provide more details and examples in \cref{app:tsne}.
The representation reminisces a Möbius strip indicating exactly the looked-after continuity.
Interestingly, such latent structure emerge without the Toeplitz prior.
We provide an profound study on the necessity of the prior in \cref{app:toeplitz_prior}.

\subsection{Generalization and varying Number of Beams}
\label{sec:receptive_field}

\begin{figure*}[ht]
\vskip 0.2in
\begin{center}
\centerline{\includegraphics[width=\textwidth]{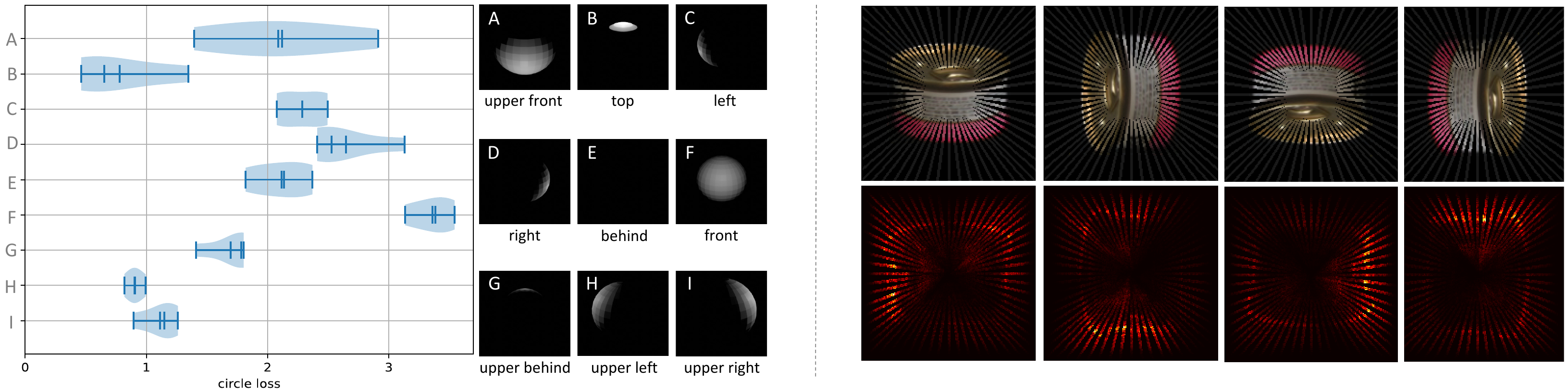}}
\caption{\textit{(left)} \textit{COIL100} pre-trained \gls*{model} tested on synthetic spheres with different light positions and \textit{(right)} Saliency maps of a \textit{COIL100} sample.}
\label{fig:light_saliency_combi}
\end{center}
\vskip -0.2in
\end{figure*}

We are interested in the generalisation capabilities of \gls*{model}.
In that light, we evaluate the variance of our model when trained and tested on the same distribution.
To go one step further, we also evaluated the test performance on all possible rotations, $\mathtt{SO(2)}$, when trained only on a subgroup.
All this is conducted using a set of different number of beams $|\mathcal{B}|$ to reason about the influence of the receptive field on the prediction quality.
We used the \textit{COIL100} dataset for our investigations using a beam length of $D = W/2 - \delta$ to reduce the risk of overfitting.
The performance results are composed in \cref{fig:experiment_plot}B.
Generally, we observe a performance improvement with increasing $|\mathcal{B}|$, which was expected.
The higher the sampling rate, the more information are available to infer the rotation angle.


Blue curves indicate the training and testing performance, respectively, on the finite group $\mathcal{C}_{|\mathcal{B}|}$.
Orange curves illustrate the performance on the $\mathtt{SO(2)}$, \textit{i.e.}, continuous angle regression.
We found that the variance of the model trained and tested on $\mathtt{SO(2)}$ is much larger than in the subgroup experiments.
This is reasonable due to the significantly larger output space.
It is surprising, however, that the training performance on $\mathtt{SO(2)}$ is lower than on $\mathcal{C}_{|\mathcal{B}|}$.
Most interesting is the red curve, which illustrates the test performance on $\mathtt{SO(2)}$, when trained on $\mathcal{C}_{|\mathcal{B}|}$.
Here, the model is forced to interpolate and can not further rely on the bijection between $k$ and $\theta$ (see \cref{eq:sym_group}).
To increase the difficulty, even more, we kept the sample complexity constant, \textit{i.e.}, the number of rotations per image in that dataset.
The result indicates that \gls*{model} can generalise to more significant distributions and even continuous regression.
Another exciting finding is that this setting outperforms the $\mathtt{SO(2)}$-trained model.
We leave both of these remarkable findings for possible future work.

\subsection{Leveraging Light Reflections} \label{sec:light}

As humans, we learn canonical orientations for many different object classes, as shown by \cite{Harris2001}.
This originates from our physical understanding of the perceived world.
For instance, our mental model of a giraffe is most likely in the form of an upstanding animal. 
What if the presented object does not come with such a bias, like a ball?
Given that the image is shooted in a room with static lightning conditions, this can be used as an indicator for its canonical orientation. 
We hypothesise that this salient feature is the main driver for the canonicalisation of such images.
To test our hypothesis, we pre-train our model on a dataset with a spatially static light, as \textit{COIL100}.
During testing, we utilise different test sets, each with a different light position.
If reflections are a salient feature to determine the canonical orientation, then the prediction performance must vary across the sets.
We simulated different lightning conditions in \cite{Blender} with a sphere as the target object to purify the scene by focusing only on reflections.
We generated $100$ sample images with different light intensities for each of the nine different light positions.
The results in \cref{fig:light_saliency_combi} \textit{(left)} show noticeable prediction performances across the test sets.
This indicates that the model trained on \textit{COIL100} indeed learned to leverage the reflections as the salient feature for the angle regression.
This finding is supported by the saliency maps in \cref{fig:light_saliency_combi} \textit{(right)}, where the model focuses on the outline and the shadow areas of the object.
The plots also show the equivariance between the input and the feature map under rotation.
If the data is recorded under static lighting conditions, as in industrial environments, reflections are a robust salient feature for adequately estimating the orientation across the train and test set.
This explains why the variance in all of our experiments is reasonably low.

\subsection{Performance Impact of Beam Length}
\label{app:beam_lengths}

\begin{figure*}[ht]
\vskip 0.2in
\begin{center}
\centerline{\includegraphics[width=\textwidth]{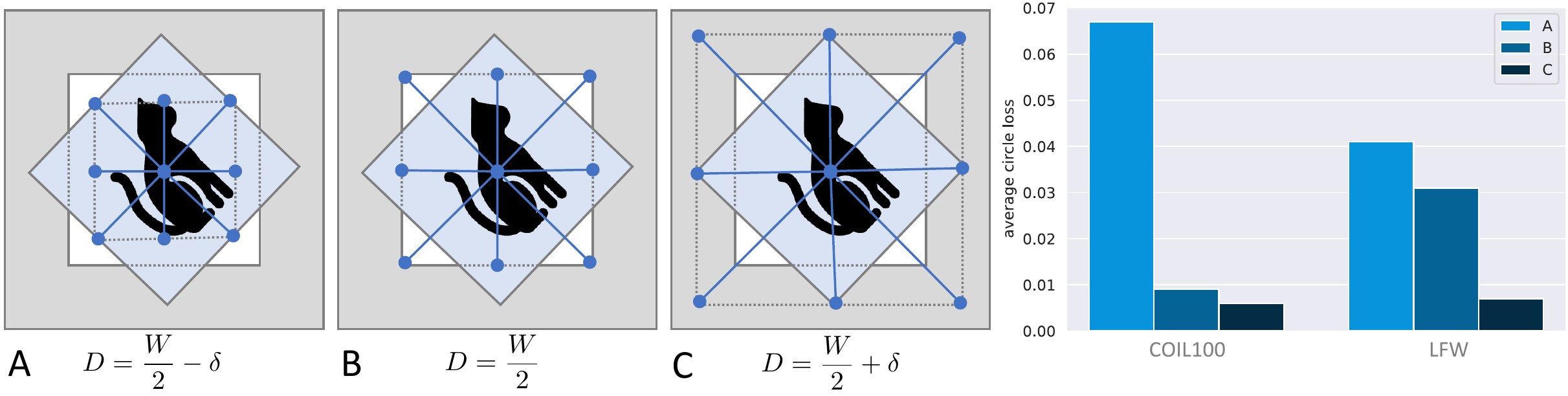}}
\caption{Superficial illustration of the three different beam lengths and its performance impact on two test sets, that is \textit{COIL100} and \textit{LFW}.}
\label{fig:beam_lengths}
\end{center}
\vskip -0.2in
\end{figure*}

During the rotation of any image, colour values move on circular tracks around the centre.
However, the image shape is rectangular rather than circular.
Therefore, pixels on outer circular tracks might not get a colour value assigned.
These rotation artifacts are visualized in \cref{fig:beam_lengths} and highlighted in white.
We call these edges between the vanilla image and any padded region (including these artefacts) image-to-padding borders.
Beams that include these regions hold valuable information for the model to overfit on this sharp colour gradient.
There are three beam lengths, as shown in \cref{fig:beam_lengths} that offer particularly interesting semantic contexts for the angle regression:
\begin{itemize}
    \item[\textbf{A}] We set $D = W/2 - \delta$ wit $\delta$ according to \cref{lemma:padding}.
    This prevents beams from covering padded pixels across all possible rotations.
    On the one hand, this causes a drastic context reduction but, on the other hand, eliminates the possibility of overfitting on the image-to-padding border.
    \item[\textbf{B}] We set $D = W/2$. Here the context is increased, but the model might overfit the included image-to-padding border.
    \item[\textbf{C}] We set $D = W/2 + \delta$. The full context is provided under all possible rotations.
    However, significant amounts of input pixels provide no information.
\end{itemize}
We conducted two experiments, where we trained and evaluated \gls*{model} on \textit{COIL100} and \textit{LFW}, respectively.
Beams with full context (\textbf{C}) achieved consistently lower $\mathcal{L}_{\text{circle}}$ loss as shown in \cref{fig:beam_lengths}.
Due to the significant loss increase under setting \textbf{A}, we hypothesise that \gls*{model} overfitted on the image-to-padding border.
To support this hypothesis, we conducted another experiment.
\\
If the target object on the image lives on a monochrome background, the image-to-padding border would be eliminated.
However, all utilised datasets, including \textit{COIL100}, do not show the object on a purely black background.
Against this background, we used an adaptive padding technique to colour the background monochrome.
We extracted the colour value of a corner pixel of the vanilla \textit{COIL100} target image.
The gathered colour code is then utilised for image-individual padding.
The saliency maps \cite{simonyan2014deep} in \cref{fig:padding} show that the model learns the two-dimensional shape of the target object instead of focusing on the image-to-padding border.


\section{Conclusion}
We presented an angle regressor for image data based on radial beams sampled from the input image.
Our \acrfull*{model} maps random centre-rotated images to their canonicalised orientations.
This allows for model-agnostic rotation-sensitive downstream prediction networks.
Our model is part of a disjointly trained vision pipeline comprising an object detection and cropping mechanism, followed by \acrfull*{model} and a downstream classificator.
Through the object detection and cropping mechanism, we ensure that the target object is centred, removing the centre bias of our radial beam sampling.
\gls*{model} achieves a regression error of $13^\circ$ on \textit{COIL100} and $12^\circ$ on \textit{LFW} with random continuous rotations.
A possible application domain is robotic handling, where a robot arm needs to orient its end-effector according to the orientation of the target object.
\gls*{model} holds the potential to supersede known but inefficient strategies, like augmented training as in \cite{Gudimella2017}.
During our investigations, our angle distance loss learned a meaningful structured latent space.
Besides empirical findings, we provide a mathematical foundation for our proposed radial beam sampling to ease future investigations.
This includes a potential study of the information decrease along beams towards the centre, \textit{i.e.}, pixels further from the centre are more affected by rotations than pixels closer to the centre.
Pixels towards the centre are represented redundantly due to the overlapping nature of beams.
Furthermore, end-to-end training using \gls*{model} as the localisation sub-network in Spatial Transformers \cite{jaderberg2016spatial} would be an exciting future direction.

\section{Acknowledgements}
The authors acknowledge the financial support by the Federal Ministry of Education and Research of
Germany (BMBF) within the framework for the funding for the project PASCAL.

\bibliography{example_paper}
\bibliographystyle{icml2023}

\newpage
\appendix
\onecolumn

\section{Graph Neural Networks}
\label{app:gnns}

\glspl*{gnn} enable learning on graphs and thus directly leverage the underlying data structure as an inductive bias.
The basis for this is a learnable inference method, \textit{i.e.} the iterative parameterized message passing procedure.
For simplicity, we consider only graph convolutional neural networks \cite{kipf2017semisupervised}, which we will use interchangeably with \glspl*{gnn} in this work.
Furthermore, we use node and vertex as synonyms and edge and link to denote connections.
At each layer $l$ the node features of each node are updated by the \gls*{gnn} propagation rule
\begin{equation} \label{eq:gcn_aggregation}
  \boldsymbol{h}^{l+1}_v \leftarrow f_{node} \left(\underset{u \in \mathcal{N}(v)}{\bigcirc} f_{link} \left(\boldsymbol{h}^{l}_v, \boldsymbol{h}^{l}_u \right), \boldsymbol{h}^{l}_v \right),
\end{equation}
where $\boldsymbol{h}^{l+1}_v$ is the updated node embedding (output of layer $l$). $\bigcirc$ denotes a permutation invariant operation, like summation, maximization, minimization, or averaging \cite{DeepSets}.
This is the reason for \gls*{gnn} being permutation invariant \textit{w.r.t.} their local node set.
Therefore, graphs convey the underlying topology, not the geometry, \textit{i.e.} distances and angles between nodes.
By this, the symmetry of graphs is encoded such that the connection of nodes is relevant and not their mere position. 
However, this is not always beneficial since it worsens the representational power and causes steerability downsides for some use cases \cite{kondor2018covariant}.
Although mean and max-pooling aggregators are well-defined multiset functions, they are not injective \cite{xu2019powerful}.
For multiset injective aggregators, this learning procedure closely resembles the well-known \cite{Weisfeiler1968} Graph Isomorphism Test.
This test answers the question of whether or not two graphs are topologically equivalent.
Note, however, that the outcome of this test provides a necessary but insufficient condition for graph isomorphisms.
As shown in \cite{xu2019powerful}, the sum operator holds the highest expressiveness among permutation invariant aggregator choices.
\\
\cite{farina2021symmetrydriven} proposed an angle preserving graph network, which encodes the direction of the neighbour nodes.
For $l=0$ we initialize $\boldsymbol{h}^{l}_v$ by the node features $\boldsymbol{x}_v$.
This update mechanism is a composition of a node embedding function $f_{node}$ and a link embedding function $f_{link}$, where both are modelled by neural networks rendering the procedure learnable.
$f_{link}$ models the message passing procedure, where the query node embedding $\boldsymbol{h}^{l}_v$ and its neighbor node embedding $\boldsymbol{h}^{l}_u$ are combined in a non-linear fashion.
This allows for an information flow from the neighbour nodes $\mathcal{N}(v)$ of query node $v$ to itself. In  \cref{eq:gcn_aggregation}, we used the sum operator as the aggregator of all messages.
The result is passed into the node embedding function $f_{node}$, modelled by a neural network, which encodes the aggregated information and the current query node embedding.
Node states are propagated until an equilibrium is obtained.
This iterative convolutional process allows for the treatment of irregular data.

\section{Proofs}

\subsection{Beam Overlaps and Coverage Approximation}
\label{app:overlap}

\begin{figure}[t!]
  \centering
  \includegraphics[scale=1.0]{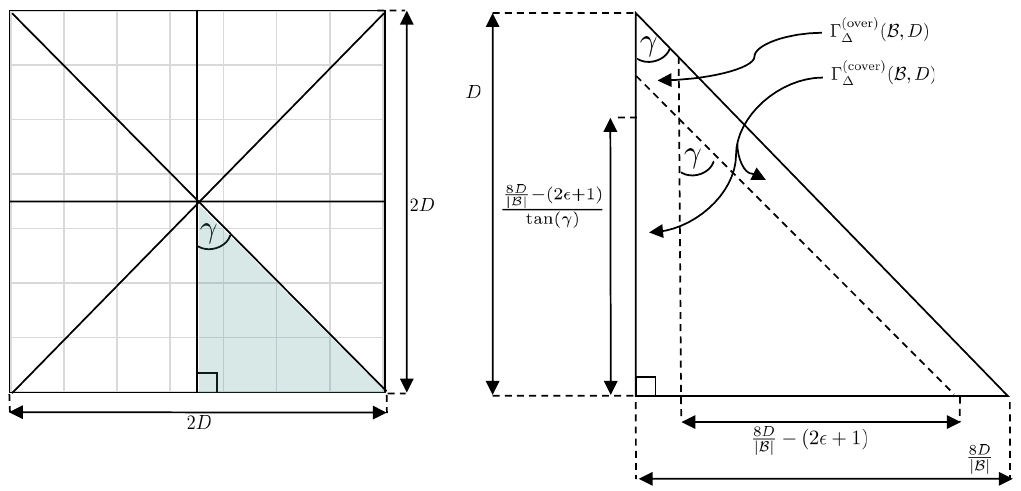}
  \caption{\textit{(left)} Beam sketch with $|\mathcal{B}| = 8$ on the pixel grid $\Omega$. 
  Each beam has a fixed length of $D$, thus a two-dimensional hull of length and height $2D$ results.
  \textit{(right)} One example triangle with base length and height annotated with the deduced formulas, as well as the coverage $\mcover(|\mathcal{B}|, D)$ and overlap $\mover(|\mathcal{B}|, D)$. Within the triangle, beam widths are depicted by dashed lines.}
  \label{fig:overlap}
\end{figure}

\begin{proof}
    We use a linear relaxation of the discrete pixel grid to quantify the coverage and overlap.
    Under the assumption of reflection symmetric beams, where we can position the symmetry axis at every beam and opposite beam pair (see  \cref{fig:overlap}).
    Therefore, the space is chopped into equally sized triangles.
    It is sufficient to estimate quantities for one triangle and extrapolate to the full scope.
    Let $\gamma = 360^\circ / |\mathcal{B}|$ be the angle of each triangle at the center of the image.
    We can leverage the spatial bound of any new beams, which is hemmed in by its adjacent neighbor beams.
    We say that a pixel of a beam either overlaps another beam's pixel or not.
    As aforementioned all beam lengths are limited to $D$ pixels.
    Let $\mcover, \mover \in \mathbb{R}_+$ be the coverage and overlap of triangle $\Delta$, respectively.
    The coverage is a strict upper bound for the overlap $\mcover > \mover$ since the last pixel of any beam needs to be unique otherwise the beam already exists in $\mathcal{B}$.
    Intuitively, the narrower the neighbors, the tighter the spatial bound, the higher the overlap.
    The base length of each triangle can be computed by
    \begin{equation}
      \frac{2D}{\frac{1}{4} |\mathcal{B}|} = \frac{8D}{|\mathcal{B}|}.
    \end{equation}
    Part of the approximation is the assumption that all are orthogonal triangles.
    Under the linear relaxation we can compute the surface area of an triangle by
    \begin{equation}
      \Delta(\mathcal{B}, D) = \frac{1}{2} \frac{8D}{|\mathcal{B}|} D = \frac{4 D^2}{|\mathcal{B}|}.
    \end{equation}
    The surface area of the inner embedded triangle, \textit{i.e.}, the uncovered area, can be computed using different trigonometric ratios.
    Here, we leverage the fact that $\gamma$ appears also as the top angle of the embedded triangle, as illustrated in \cref{fig:overlap}.
    To estimate the uncovered surface area, we need width of beams, which is given by $(2\epsilon + 1)$ with $\epsilon$ being the thickness.
    Then, the base length of this inner triangle is
    \begin{equation}
      \frac{8D}{|\mathcal{B}|} - (2\epsilon + 1).
    \end{equation}
    Using a trigonometric ratio for orthogonal triangles the height of the triangle can be computed by
    \begin{equation}
        \frac{\frac{8D}{|\mathcal{B}|} - (2\epsilon + 1)}{\tan(\gamma)}.
    \end{equation}
    Then, the uncovered surface area is approximately
    \begin{equation}
          \Delta^\prime(|\mathcal{B}|, D) \approx
          \frac{1}{2} \frac{\frac{8D}{|\mathcal{B}|} - (2\epsilon + 1)}{\tan(\gamma)} \left( \frac{8D}{|\mathcal{B}|} - (2\epsilon + 1) \right)
          \approx \frac{\left( \frac{8D}{|\mathcal{B}|} - (2\epsilon + 1) \right)^2}{2\tan(\gamma)}.
    \end{equation}
    Adding together the beam thickness and the uncovered area and subtracting the overall surface area gives the overlap
    \begin{equation}
      \mover(|\mathcal{B}|, D) \approx (2\epsilon + 1)D + \Delta^\prime(|\mathcal{B}|, D) - \Delta(|\mathcal{B}|, D).
    \end{equation}
    This allows us to compute the coverage by subtracting this overlap from the beam thickness
    \begin{equation}
      \mcover(|\mathcal{B}|, D) \approx (2\epsilon + 1)D - \mover(|\mathcal{B}|, D).
    \end{equation}
    Due to the linear relaxation and the partitioning of the area in equivalent triangles, a linear extrapolation to the full image is possible.
    Such that, the total overlap and coverage are given by
    \begin{align}
          \Gamma_{\text{over}}(|\mathcal{B}|, D) &\approx |\mathcal{B}| \mover(|\mathcal{B}|, D) \\
          &\approx |\mathcal{B}| \left[ (2\epsilon + 1)D + \Delta^\prime(|\mathcal{B}|, D) - \Delta(|\mathcal{B}|, D) \right] \\
          &\approx |\mathcal{B}| \left[ (2\epsilon + 1)D + \frac{\left( \frac{8D}{|\mathcal{B}|} - (2\epsilon + 1) \right)^2}{2\tan(\gamma)} - \frac{4 D^2}{|\mathcal{B}|} \right],
    \end{align}
    \begin{align}
      \Gamma_{\text{cover}}(|\mathcal{B}|, D) &\approx |\mathcal{B}| \mcover(|\mathcal{B}|, D) \\
      &\approx |\mathcal{B}| \left[ (2\epsilon + 1)D - (2\epsilon + 1)D - \Delta^\prime(|\mathcal{B}|, D) + \Delta(|\mathcal{B}|, D) \right] \\
      &\approx |\mathcal{B}| \left[ \frac{4 D^2}{|\mathcal{B}|} - \frac{\left( \frac{8D}{|\mathcal{B}|} - (2\epsilon + 1) \right)^2}{2\tan(\gamma)} \right].
    \end{align}
Both approximations ground on the assumption of triangles and thus subject to $|\mathcal{B}| \geq 8$.
\end{proof}


\subsection{Maximal Beam Set Cardinality Upper Bound}
\label{app:beam_upper_bound}

\begin{proof}
  Each beam needs to cover at least one pixel not covered by others, otherwise these beams would be redundant.
  We say that the maximum number of beams $|\mathcal{B}_{max}|$ is reached if the receptive field fully covers the image.
  Therefore, we use the coverage approximation from \cref{lemma:cover}
    \begin{align}
      \Gamma_{\text{cover}}(|\mathcal{B}|, D) &\approx |\mathcal{B}| \left[ \frac{4 D^2}{|\mathcal{B}|} - \frac{\left( \frac{8D}{|\mathcal{B}|} - (2\epsilon + 1) \right)^2}{2\tan(\gamma)} \right].
    \end{align}
  To obtain $|\mathcal{B}_{max}|$ we compute the derivation \textit{w.r.t.} $|\mathcal{B}|$, such that
    \begin{align}
      \frac{\partial\Gamma_{\text{cover}}(|\mathcal{B}, D)}{\partial|\mathcal{B}|} &= \frac{\frac{64D^2}{|\mathcal{B}|^2} - (2\epsilon + 1)^2}{2\tan\gamma} = 0 \\
      |\mathcal{B}| &= \sqrt{\frac{64D^2}{(2\epsilon+1)^2}} = \frac{8D}{(2\epsilon+1)} \geq |\mathcal{B}_{max}|.
    \end{align}
  Due to the violated integer constraint this defines an upper bound for $|\mathcal{B}_{max}|$.
\end{proof}

\subsection{Optimal Padding Margin}
\label{app:padding}

\begin{proof}
  Let $W \in \mathbb{N}$ and $H \in \mathbb{N}$ denote the width and height, respectively, of an squared image $X$ with $W = H$.
  This sets the center point at $\boldsymbol{c} = [W / 2, W / 2]$.
  The image can be zero padded uniformly at the borders by a margin $\delta \in \mathbb{N}$.
  The goal is to find $\delta$ such that any arbitrary rotation of the image around its center will not cause any information loss.
  Information is lost if any pixel of $X$, which is bounded spatially by the unpadded vanilla image region, gets rejected during rotation.
  Using a rotation matrix $R \in [1, -1]^4$ pixel vectors are mapped to real vectors.
  When enforcing the rigid discrete pixel grid, pixels might get cut off.
  Let $\boldsymbol{d}$ be the maximal distant pixel from $\boldsymbol{c}$, which clearly lies in the corners of the image.
  By the Pythagorean theorem we have $||\boldsymbol{d}||_2 = \sqrt{(W / 2)^2 + (H / 2)^2} = \sqrt{2}(W / 2)$ using the assumption above.
  The cut off is maximal if the image is rotated by $45^\circ$ and hence $\boldsymbol{d}$ is orthogonal to the width and height axes.
  So, the required padding can be quantified by the ceiled offset between $||\boldsymbol{d}||_2$ and $W / 2$ such that
  \begin{equation}
      \delta = \lceil{\max \left(0, \left\lceil ||\boldsymbol{d}||_2 - \frac{W}{2} \right\rceil \right)}
    = \max \left(0, \left\lceil \sqrt{2}\frac{W}{2} - \frac{W}{2} \right\rceil \right)
    = \max \left(0, \left\lceil \frac{1}{2} W \left(\sqrt{2} - 1 \right) \right\rceil \right).
  \end{equation}
\end{proof}

\section{Loss Analysis}
\label{app:loss}

\subsection{Circle Loss preserves Angle Distances}
\label{app:circle_loss}

One-dimensional real values lie on the number line $\mathbb{R}$ or on a subset of it if bounds apply.
This renders distance measuring trivial, \textit{e.g.} we might use the absolute difference between two scalars.
Angles, on the other hand, lie on a circle, such that $0^\circ$ and $359^\circ$ are closer as $0^\circ$ and $2^\circ$.
We show that our circle loss 
\begin{equation}
\nonumber
    \mathcal{L}_{\text {circle }}(\theta, \boldsymbol{z})=(\sin (\theta)-i m(\boldsymbol{z}))^{2}+(\cos (\theta)-r e(\boldsymbol{z}))^{2}
\end{equation}
preserves angle distances and is therefore highly expressive.
It is well known, that the minimum and maximum numerical angle values are $0^\circ$ and $360^\circ$, respectively.
Let $d(\theta, \theta^\prime) \in \mathbb{R}_+$ be a distance measure between two arbitrary angles $\theta$ and $\theta^\prime$.
We seek a distance measure where the following two conditions holds
\begin{equation}
    \lim_{\theta \rightarrow 360^\circ} d(\theta, 0^\circ) = 0,
\end{equation}
\begin{equation}
    \lim_{\theta \rightarrow 0^\circ} d(\theta, 360^\circ) = 0.
\end{equation}
Using our circle loss it is easy to show that both conditions hold, \textit{i.e.}
\begin{equation}
    \begin{aligned}
        \lim_{\theta \rightarrow 360^\circ} &(\sin (0^\circ)- \sin (\theta) )^{2}+(\cos (0^\circ)-\cos (\theta))^{2} \\
        = &(\sin (0^\circ)- \sin (360^\circ) )^{2}+(\cos (0^\circ)-\cos (360^\circ))^{2} = 0,
    \end{aligned}
\end{equation}
\begin{equation}
    \begin{aligned}
        \lim_{\theta \rightarrow 0^\circ} &(\sin (360^\circ)- \sin (\theta) )^{2}+(\cos (360^\circ)-\cos (\theta))^{2} \\
        = &(\sin (360^\circ)- \sin (0^\circ) )^{2}+(\cos (360^\circ)-\cos (0^\circ))^{2} = 0.
    \end{aligned}
\end{equation}
Therefore, we conclude that our circle loss indeed respects the angle distances.

\subsection{Extrema Analysis}
\label{app:extrema}

\begin{figure}[h!]
  \centering
  \includegraphics[scale=0.45]{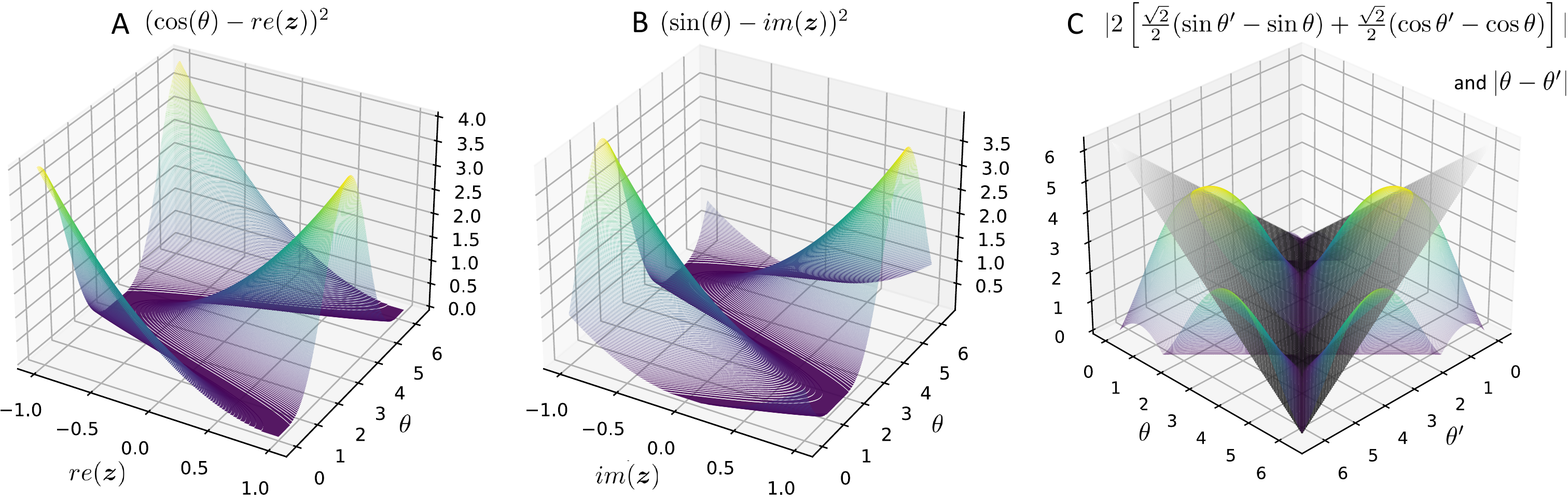}
  \caption{\textbf{A} and \textbf{B} illustrate both loss term surface plots, while \textbf{C} depicts the plot of both Lipschitz relevant inequality terms with $im(\boldsymbol{z}), re(\boldsymbol{z}) = \sqrt{2} / 2$.}
  \label{fig:loss_surface}
\end{figure}


For an extrema analyis we first derive the partial derivatives of the multivariable loss
\begin{equation}
\nonumber
    \mathcal{L}_{\text{circle}}(\theta, \boldsymbol{z}) = (\sin(\theta) - im(\boldsymbol{z}))^2 + (\cos(\theta) - re(\boldsymbol{z}))^2.
\end{equation}
The input lies in the cubic domain bounded by $\hat{\theta} \in [0, 2\pi)$ and $\boldsymbol{z} \in [-1, 1]^2$.
The gradient is given by
\begin{equation}
  \nabla \mathcal{L}_{\text{circle}}(\theta, \boldsymbol{z}) = \left[
  \frac{\partial \mathcal{L}_{\text{circle}}(\theta, \boldsymbol{z})}{\partial \theta},
  \frac{\partial \mathcal{L}_{\text{circle}}(\theta, \boldsymbol{z})}{\partial re(\boldsymbol{z})},
  \frac{\partial \mathcal{L}_{\text{circle}}(\theta, \boldsymbol{z})}{\partial im(\boldsymbol{z})}
  \right]^\top.
\end{equation}
This resolves in
\begin{equation}
  \label{eq:grad_loss_beta}
  \begin{aligned}
    \frac{\partial}{\partial \theta} \mathcal{L}_{\text{circle}}(\theta, \boldsymbol{z})
    &= 2\cos(\theta)(\sin(\theta) - im(\boldsymbol{z}))
    - 2\sin(\theta)(\cos(\theta) - re(\boldsymbol{z})) \\
    &= 2\left[ re(\boldsymbol{z})\sin(\theta) - im(\boldsymbol{z})\cos(\theta) \right],
  \end{aligned}
\end{equation}
\begin{equation}
  \label{eq:grad_loss_re}
  \frac{\partial}{\partial re(\boldsymbol{z})} \mathcal{L}_{\text{circle}}(\theta, \boldsymbol{z})
  = 2(\cos(\theta) - re(\boldsymbol{z})),
\end{equation}
\begin{equation}
  \label{eq:grad_loss_im}
  \frac{\partial}{\partial im(\boldsymbol{z})} \mathcal{L}_{\text{circle}}(\theta, \boldsymbol{z})
  = 2(\sin(\theta) - im(\boldsymbol{z})).
\end{equation}
Analysing the behaviour of  \cref{eq:grad_loss_re} and \cref{eq:grad_loss_im} at zero indicates that a critical function is given at $im(\boldsymbol{z}) = \sin(\theta)$ and $re(\boldsymbol{z}) = \cos(\theta)$.
We provided a visual support in \cref{fig:loss_surface}, where we plotted both loss terms of \cref{eq:circle_loss}.
The minimas lie along $\sin$ and $\cos$.
Intuitively, due to the squares in  \cref{eq:circle_loss} and the sum operation as loss term concatenation, there is no room for other minimas.
Since predictions and ground truth values have numerical bounds, the loss is upper bounded as well.
To quantify this upper bound, we aim to solve
\begin{equation}
   \theta^\ast, \boldsymbol{z}^\ast = \argmax_{\theta, \boldsymbol{z}} \text{ } \mathcal{L}_{\text{circle}}(\theta, \boldsymbol{z}).
\end{equation}
We maximize the loss by using diametrical unit vector predictions compared to the ground truth, \textit{i.e.}, $im(\boldsymbol{z})^\ast = \cos(\theta)$ and $re(\boldsymbol{z})^\ast = \sin(\theta)$.
This leads to
\begin{equation}
   \theta^\ast = \argmax_{\theta} (\sin(\theta) - \cos(\theta))^2 + (\cos(\theta) - \sin(\theta))^2 = \frac{3}{4}\pi.
\end{equation}
So, we have the upper bound $\mathcal{L}_{\text{circle}}(\frac{3}{4}\pi, [\sin(\theta^\ast), \cos(\theta^\ast)]) = 4$.

\subsection{Lipschitz Smoothness}
\label{app:lipschitz}

Our loss function is Lipschitz smooth \textit{iff} the following condition hold
\begin{equation}
  \| \mathcal{L}_{\text{circle}}(\theta, \boldsymbol{z}) - \mathcal{L}_{\text{circle}}(\theta^\prime, \boldsymbol{z}) \| \leq K \| \theta - \theta^\prime \|,
\end{equation}
where $K$ is the Lipschitz constant.
Using the definition of the loss in  \cref{eq:circle_loss} gives
\begin{equation}
  \begin{aligned}
    \| &(\sin(\theta) - im(\boldsymbol{z}))^2 + (\cos(\theta) - re(\boldsymbol{z}))^2 - \\
    &(\sin(\theta^\prime) - im(\boldsymbol{z}))^2 - (\cos(\theta^\prime) - re(\boldsymbol{z}))^2 \| \leq K \| \theta - \theta^\prime \|.
  \end{aligned}
\end{equation}
This dissolves in
\begin{equation}
  \begin{aligned}
    \| &\sin^2(\theta) - 2im(\boldsymbol{z})\sin(\theta) + im(\boldsymbol{z})^2 + \\
    &\cos^2(\theta) - 2re(\boldsymbol{z})\cos(\theta) + re(\boldsymbol{z})^2 - \\
    &\sin^2(\theta^\prime) + 2im(\boldsymbol{z})\sin(\theta^\prime) - im(\boldsymbol{z})^2 - \\
    &\cos^2(\theta^\prime) + 2re(\boldsymbol{z})\cos(\theta^\prime) - re(\boldsymbol{z})^2 \|
    \leq K \| \theta - \theta^\prime \|.
  \end{aligned}
\end{equation}
Using the trigonometrical identity $\sin^2(\theta) + \cos^2(\theta) = 1$ and some rearrangements give
\begin{equation}
    \| 2 \left[ im(\boldsymbol{z}) (\sin(\theta^\prime) - \sin(\theta)) + re(\boldsymbol{z}) (\cos(\theta^\prime) - \cos(\theta)) \right] \|
    \leq K \| \theta - \theta^\prime \|.
\end{equation}
Assuming no trivial case, \textit{i.e.}, $\theta \neq \theta^\prime$, the projection on the unit circle gives always smaller norms
\begin{equation}
    \| \sin(\theta^\prime) - \sin(\theta) \| < \| \theta - \theta^\prime \|,
\end{equation}
\begin{equation}
    \| \cos(\theta^\prime) - \cos(\theta) \| < \| \theta - \theta^\prime \|.
\end{equation}
Furthermore, per definition we have $\| \boldsymbol{z} \| = 1$, thus only a fraction of both terms above are aggregated.
With $K=2$ the Lipschitz inequality can be guaranteed.
In  \cref{fig:loss_surface}C we plotted both functions of the inequality.

\section{Implementation Details and Further Experiments}

\begin{figure*}[ht]
\vskip 0.2in
\begin{center}
\centerline{\includegraphics[width=\textwidth]{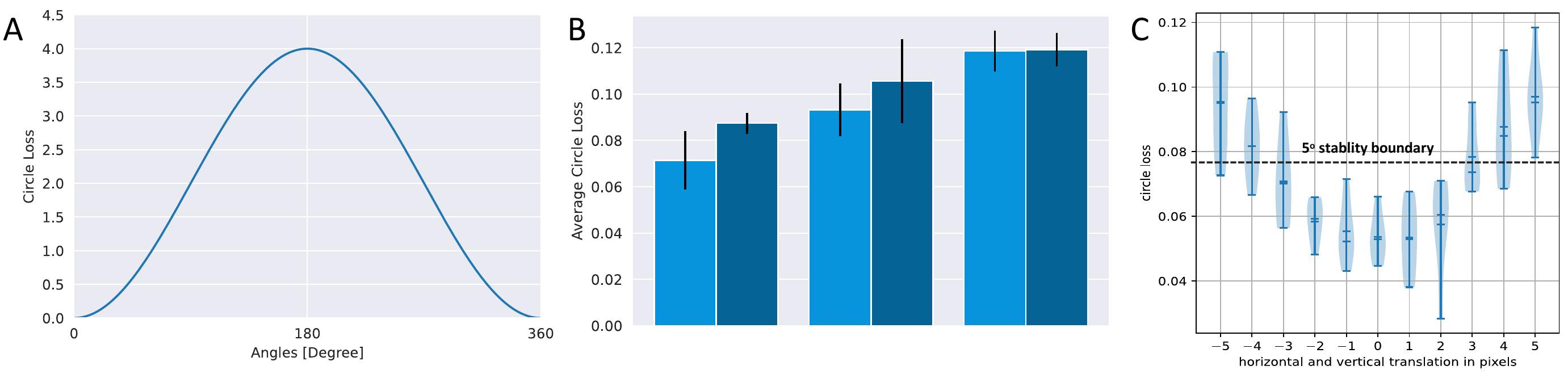}}
\caption{\textit{(A)} Relationship between the circle loss $\mathcal{L}_{\text{circle}}$ and angles in degree. 
\textit{(B)} Training and test performance using three different losses over five runs.
\textit{(C)} Translation robustness on \textit{COIL100}.}
\label{fig:experiment_plot2}
\end{center}
\vskip -0.2in
\end{figure*}

For a better understanding of the solution quality of our angle regressor, we illustrated the non-linear relationship between circle loss $\mathcal{L}_{\text{circle}}$ values and angles in degree in \cref{fig:experiment_plot2}A.
For example, if we rotate an image by $180^\circ$, the difference between both images can also be quantified by $\mathcal{L}_{\text{circle}} = 4$.
With this in mind, we first provide more implementation details of our basic experimental setup, followed by further experiments.

\subsection{Base Model Specification}
\label{sec:implementation_details}

The beam encoder is partitioned into a proximity encoder and a spatial encoder.
The proximity encoder uses 2D convolutional kernels shifting over each beam and compresses the signal down to a latent subspace $\mathbb{R}^{|\mathcal{B}| \times 2\epsilon + 1 \times D \times C} \rightarrow \mathbb{R}^{|\mathcal{B}| \times D-8 \times L/8}$.
We use valid padding for all convolutions to reduce the dimensionality.
The subsequent beam encoder processes the embedding with 1D spatial convolution $\mathbb{R}^{|\mathcal{B}| \times D-8 \times L/8} \rightarrow \mathbb{R}^{|\mathcal{B}| \times L}$.
We use a $128$-dimensional latent space (\textit{i.e.}, $L=128$).
The context and neighbourhood information among beams is encoded using a three-layer \gls*{gnn}.
The obtained embeddings and their ordering are decoded in a three-layer \gls*{lstm}, representing the permutation decoder $\mathbb{R}^{|\mathcal{B}| \times L} \rightarrow \mathbb{R}^{L}$.
We utilize \textit{LeakyReLU} activations as piece-wise non-linearities, \textit{i.e.}, $\max(0.3 x, 0)$, which empirically found to be more performant than \textit{ReLUs}.
Weight initialisation according to \cite{HeInit} is used with a bias initialisation close to $0$ to accompany the rectified network well.
Eventually, three linear layers and a subsequent normalisation transform the final latent state of the permutation decoder into a complex vector $\mathbb{R}^{L} \rightarrow [-1,1]^{2}$.

We use mini-batches of $128$ elements for training with $|\mathcal{B}| = 32$ beams per image and a thickness of $\epsilon = 1$.
We use a split of $80\%$ training and $20\%$ test data.
Each dataset is augmented by randomly centre-rotated replicas of the original data points.
To this end, we sample $\theta \sim \mathcal{U}\left(\left\{360^\circ |\mathcal{B}|^{-1}k : k \in \{0, 1, \ldots, |\mathcal{B}| - 1\} \right\}\right)$.
As we will show in the following, our model can generalise well to the test data. 
Therefore, \textit{no} explicit regularisation is utilised, like dropout or specific loss terms.
We use the popular Adam optimizer \cite{Kingma2015AdamAM} with a static learning rate of $0.0001$, momentum terms $\beta_1 = 0.9$ and $\beta_2 = 0.999$ to update the model parameters.

\subsubsection{Beam Encoder}
\label{app:beam_encoder}

We partitioned the beam encoder into proximity and spatial encoder to have a clean, semantical separation.
The proximity encoder's receptive field covers the pixels' local spatial neighbourhood.
The proximity is compressed $\mathbb{R}^{2\epsilon + 1 \times D \times C} \rightarrow \mathbb{R}^{D-8 \times L/8}$ by two-dimensional convolutional kernels.
Find the layer specifications in \cref{tab:proxenc}.
Then, the remaining spatial pixel information is encoded by one-dimensional convolutional kernels, mapping $\mathbb{R}^{D-8 \times L/8} \rightarrow \mathbb{R}^{L}$.
Due to the compression, the architecture of the spatial encoder is sensitive to $D$.
We provide details of such encoders in \cref{tab:spatenc_28}, \cref{tab:spatenc_32}, \cref{tab:spatenc_128}, and \cref{tab:spatenc_250}.
We aim to preserve the spatial ordering of features for all layers, and hence no explicit pooling layers are used.
This originates from the fact that for rotational perturbation detection, the further pixels are from the centre, the more information they convey.
Pixels close to the centre not only hold redundancies due to the overlaps but also follow smaller rotation circles, and hence pixels are less sensitive to minor rotations.

\begin{table}[h!]
 \caption{The proximity encoder for $\epsilon = 1$ and  with $D>8$.}
  \centering
  \begin{tabular}{cccccc}
    \toprule
    Layer & Kernel & Padding & Strides & Non-Linearity & Feature Maps \\
    \midrule
    1 & (3, 3) & no & 1 & LeakyReLU & L/8 \\
    \bottomrule
  \end{tabular}
  \label{tab:proxenc}
\end{table}

\begin{table}[h!]
 \caption{The spatial encoder for $28 \times 28$ images with $D=14$, \textit{e.g.}, for \textit{FashionMNIST}.}
  \centering
  \begin{tabular}{cccccc}
    \toprule
    Layer & Kernel & Padding & Strides & Non-Linearity & Feature Maps \\
    \midrule
    1 & 4 & no & 1 & LeakyReLU & L/4 \\
    2 & 4 & no & 1 & LeakyReLU & L/2 \\
    3 & 4 & no & 1 & LeakyReLU & L/2 \\
    4 & 3 & no & 1 & LeakyReLU & L \\
    \bottomrule
  \end{tabular}
  \label{tab:spatenc_28}
\end{table}

\begin{table}[h!]
 \caption{The spatial encoder for $32 \times 32$ images with $D=16$, \textit{e.g.}, for \textit{CIFAR10}.}
  \centering
  \begin{tabular}{cccccc}
    \toprule
    Layer & Kernel & Padding & Strides & Non-Linearity & Feature Maps \\
    \midrule
    1 & 4 & no & 1 & LeakyReLU & L/4 \\
    2 & 4 & no & 1 & LeakyReLU & L/2 \\
    3 & 4 & no & 1 & LeakyReLU & L/2 \\
    4 & 4 & no & 1 & LeakyReLU & L \\
    5 & 2 & no & 1 & LeakyReLU & L \\
    \bottomrule
  \end{tabular}
  \label{tab:spatenc_32}
\end{table}

\begin{table}[h!]
 \caption{The spatial encoder for $128 \times 128$ images with $D=64$, \textit{e.g.}, for \textit{COIL100}.}
  \centering
  \begin{tabular}{cccccc}
    \toprule
    Layer & Kernel & Padding & Strides & Non-Linearity & Feature Maps \\
    \midrule
    1 & 5 & no & 2 & LeakyReLU & L/4 \\
    2 & 4 & no & 2 & LeakyReLU & L/4 \\
    3 & 4 & no & 1 & LeakyReLU & L/2 \\
    4 & 4 & no & 1 & LeakyReLU & L/2 \\
    5 & 4 & no & 1 & LeakyReLU & L/2 \\
    6 & 3 & no & 1 & LeakyReLU & L \\
    7 & 2 & no & 1 & LeakyReLU & L \\
    \bottomrule
  \end{tabular}
  \label{tab:spatenc_128}
\end{table}

\begin{table}[h!]
 \caption{The spatial encoder for $250 \times 250$ images with $D=125$, \textit{e.g.}, for \textit{LFW}.}
  \centering
  \begin{tabular}{cccccc}
    \toprule
    Layer & Kernel & Padding & Strides & Non-Linearity & Feature Maps \\
    \midrule
    1 & 4 & no & 2 & LeakyReLU & L/4 \\
    2 & 3 & no & 2 & LeakyReLU & L/4 \\
    3 & 4 & no & 2 & LeakyReLU & L/4 \\
    4 & 4 & no & 1 & LeakyReLU & L/2 \\
    5 & 4 & no & 1 & LeakyReLU & L/2 \\
    6 & 4 & no & 1 & LeakyReLU & L/2 \\
    7 & 3 & no & 1 & LeakyReLU & L \\
    8 & 2 & no & 1 & LeakyReLU & L \\
    \bottomrule
  \end{tabular}
  \label{tab:spatenc_250}
\end{table}

\subsubsection{Context Encoder}
\label{app:context_encoder}

We build upon the fundamentals provided in \cref{sec:methodology}.
The graph topology allows for controllable and scalable information exchange between neighbours controlled by the number of layers, \textit{i.e.}, the number of hops.
Since the ordering of neighbours matters in our case, directed edges are used to circumvent the permutation invariance of \glspl*{gnn} \cite{kipf2017semisupervised, kondor2018covariant, xu2019powerful}.
For simplicity sake, we update nodes by $\lambda Af(\mathcal{B})W$, where $f(\mathcal{B})$ is the output of the beam encoder, $A \in \{0,1\}^{|\mathcal{B}| \times |\mathcal{B}|}$ is the binary adjacency matrix and $W \in \mathbb{R}^{L \times L}$ is the learnable weight matrix.
To limit the impact of neighbour information, we use $\lambda \in (0, 1]$ as a global edge factor.
For the proposed directed wheel graph, we have an adjacency matrix
\begin{equation}
  A =
  \left[ {\begin{array}{cccccc}
    0 & 1 & 0 & 0 & \cdots & 1 \\
    0 & 0 & 1 & 0 & \cdots & 1 \\
    0 & 0 & 0 & 1 & \cdots & 1 \\
    0 & 0 & 0 & 0 & \cdots & 1 \\
    \vdots & \vdots & \vdots & \vdots & \ddots & \vdots \\
    1 & 0 & 0 & 0 & \cdots & 1 \\
  \end{array} } \right].
\end{equation}

\subsection{Toeplitz Prior Evaluation}
\label{app:toeplitz_prior}

Surprisingly, during initial empirical studies, we found that the model is learning similar latent structures and achieves better performances without the Toeplitz prior.
We illustrate our findings in \cref{fig:experiment_plot2}B, which shows the mean and standard deviation over five runs using three other losses.
The following losses are considered:
only the circle loss, $\mathcal{L} = \mathcal{L}_{\text{circle}}$,
a dynamic linear combination \textit{w.r.t.} epoch $e$, $\mathcal{L} = \left( 1 - \frac{1}{e} \right) \mathcal{L}_{\text{circle}} + \frac{1}{e} \mathcal{L}_{\text{prior}}$,
both losses summed without any scaling, $\mathcal{L} = \mathcal{L}_{\text{circle}} + \mathcal{L}_{\text{prior}}$.
As initially stated, training without the prior achieves the best results.
This might originates from either a bug in our code or the sensibility of the prior to perturbations.
That is, during the computation of the similarity matrix, we assume that there exists a beam pair combination which is nearly identical.
We formulate possible errors introduced during the rotation procedure in \cref{sec:methodology}.
If these errors are too significant, the prior might provide an erroneous momentum to the learning, which results in mediocre results.

\subsection{Downstream Evaluation}
\label{app:downstream}

We used the rotation-subset of \textit{siscore} dataset by \cite{djolonga2020robustness}, which comprises $39,540$ images of $500 \times 500 \times 3$.
Each image shows a randomly rotated object, like a truck, on a random background.
The rotation angle was sampled from $\{ 1 + (360^\circ k) / n \mid k \in [0, 1, \ldots, n-1] \}$ with $n=18$.
To match the ImageNet \cite{deng2009imagenet} shape, each image is downsampled to $224 \times 224 \times 3$.
We normalized all color values from $[0, 255]$ to $[0, 1]$ and split the dataset in $80\%$ for training and $20\%$ for testing.
Due to the insignificance of the background for the classification, no padding is added and the beam length reduced to $D=91$.
We trained \gls*{model} for $8192$ iterations on the training set and achieved a training performance of $\mathcal{L}_{\text{circle}} = 0.378$ and a test performance of $\mathcal{L}_{\text{circle}} = 0.492$.
That implies that for unseen data the prediction error is on average $41^\circ$.
This is the reason why the classification performance is not on the level of non-rotated images.
We argue, however, that with some hyper-parameter tuning this error rate can be significantly reduced.

\subsection{Dimensionality Reduction with \textit{t-SNE}}
\label{app:tsne}

\begin{figure}[t!]
  \centering
  \includegraphics[width=\textwidth]{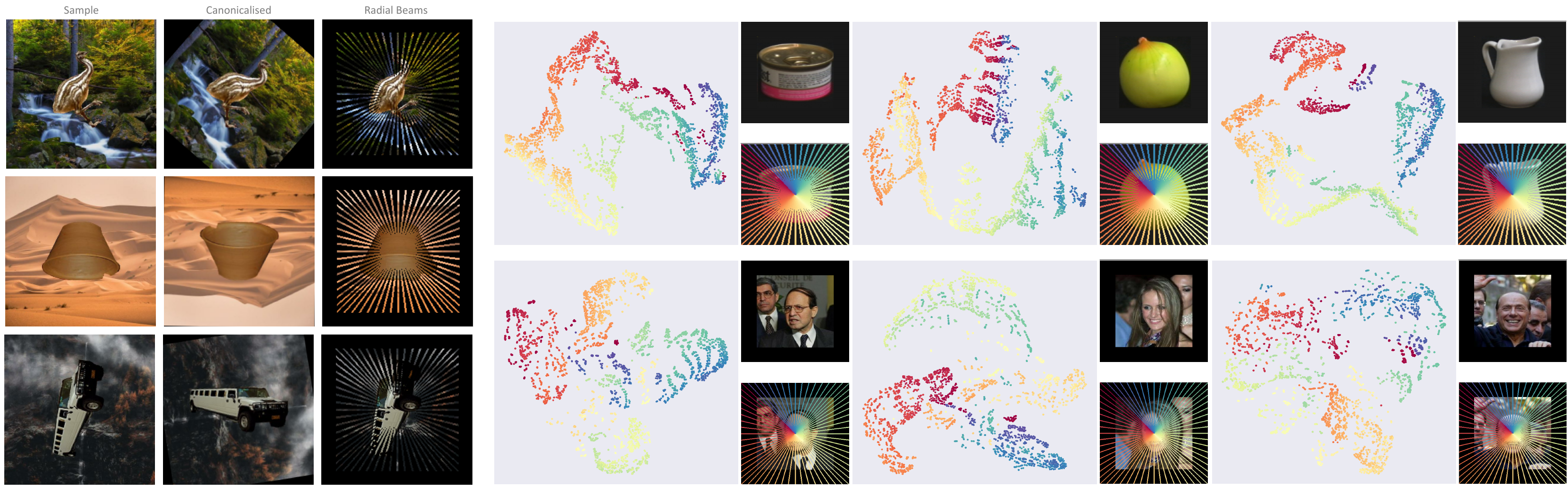}
  \caption{\textit{(left)} \textit{siscore} samples. \textit{(right)} \gls*{tsne} projections for \textit{COIL100} and \textit{LWF}.}
  \label{fig:sample_collection}
\end{figure}

\begin{figure}[t!]
    \centering
    \includegraphics[width=1.0\textwidth]{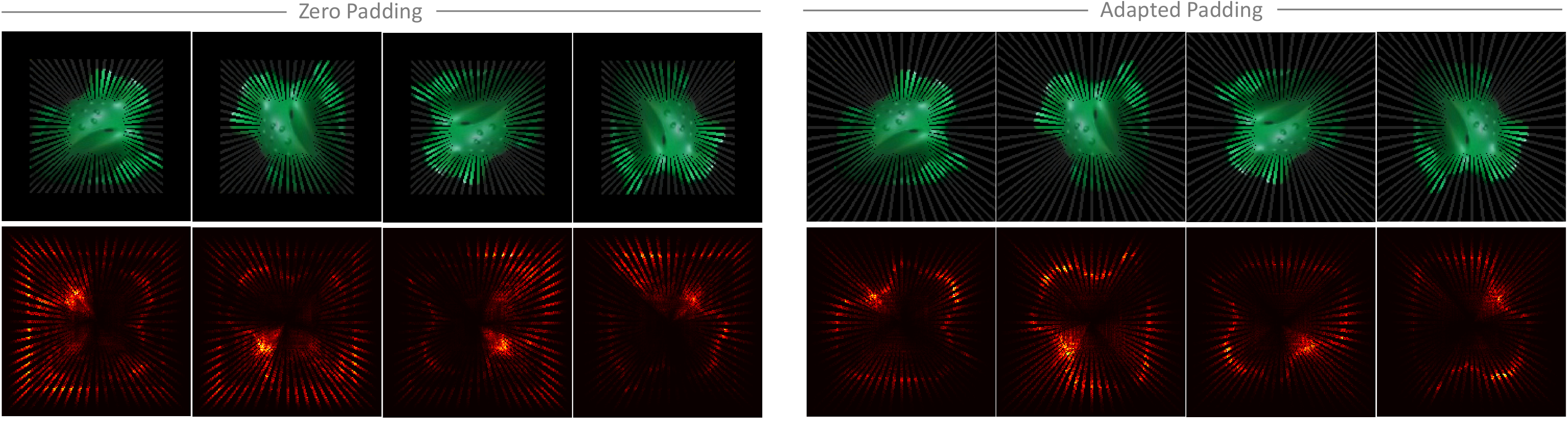}
    \caption{Saliency maps of \textit{COIL100} test sample in four rotations. Trained with standard zero padding \textit{(left)} and adapted padding \textit{(right)}.}
    \label{fig:padding}
\end{figure}

For dimensionality reduction we utilize \acrfull*{tsne}.
We use the \texttt{sklearn} implementation of \gls*{tsne} with an Barnes-Hut approximation for $1000$ iterations.
We use a random initialization, a perplexity of $100$, an early exaggeration of $12$ and the Euclidean metric.
An automatic learning rate $\rho$ is used, such that $\rho = \max(n / (0.25 e), 50)$, where $n$ is the sample size and $e$ the early exaggeration value.
Further, a minimum gradient norm of $1e-7$ for early stopping is utilized.
We trained our \gls*{model} model using the setup given in  \cref{sec:experiments}.
The model was trained on \textit{COIL100} using $\mathcal{C}_{|\mathcal{B}|}$ without the Toeplitz prior.
We used the beam embeddings from the beam encoder without the context integration to avoid bias.
A spectral colour schema to illustrate the $|\mathcal{B}|=64$ beams is employed for visual purposes.
Results are illustrated in \cref{fig:sample_collection}\textit{(right)}.
These plots show that beam embeddings sampled from different rotations of an input image form a continuous structure in the latent space.

\subsection{Geometric Stability to Translations}
\label{sec:stability}

In a real-world setting, small translations might perturb input images. 
\gls*{model} is not translation equivariant due to the radial sampling radiating from the spatially fixed centre point.
We view \gls*{model} as a part of a more extensive computer vision pipeline, where other components might compensate for this drawback.
However, our model has to fend off these small perturbations for a robust machine vision pipeline since exact upstream invariances are unlikely in real-world settings.
Theoretically, geometric stability to signal deformations is measured by a continuous complexity measure as in \cite{bronstein2021geometric}.
In this analysis, we use a subgroup of the translation group $T(2)$, which contains transformations respecting the discrete pixel grid, \textit{i.e.}, $T(2, \Omega) \subset T(2)$. 
The violin plot in \cref{fig:experiment_plot2}C shows the test performance decrease when translating the query images horizontally and vertically.
Angle predictions below the $5^\circ$ stability boundary are at most $5^\circ$ off from the unperturbed predictions.
Considering the mean of the predictions, we conclude that \gls*{model} is geometrically stable up to $\pm 3$ pixels.
This can be improved by an increased thickness $\epsilon$, which we have set to $1$ for our experiments.

\end{document}